\documentclass[10pt,journal,compsoc]{IEEEtran}
\usepackage{tabularx, booktabs, colortbl}

\newcolumntype{Y}{>{\centering\arraybackslash}X} % centered version of "X"
\newcolumntype{Z}{>{\centering\arraybackslash}p{4cm}} % modify 4cm accordingly

\usepackage{amsmath,amssymb,amsfonts}
\usepackage{algorithm}
\usepackage{algorithmic}
\usepackage{makecell}
\usepackage{threeparttable}
\usepackage{graphicx}
\usepackage{textcomp}
\usepackage{xcolor}
\usepackage{multirow}
\usepackage{dsfont}
\usepackage{booktabs}
\usepackage{hyperref}
\usepackage{amsmath, amssymb, amsfonts}
\usepackage{dsfont}

\usepackage{arydshln}

\ifCLASSOPTIONcompsoc
  \usepackage[nocompress]{cite}
\else
  \usepackage{cite}
\fi
\ifCLASSINFOpdf
\else
\fi
\begin{document}
\title{Bi-CRCL: Bidirectional Conservative-Radical Complementary Learning with Pre-trained Foundation Models for Class-incremental Medical Image Analysis}


\author{Xinyao Wu*, Zhe Xu*\textsuperscript{\dag}, Cheng Chen, Jiawei Ma, Yefeng Zheng, \IEEEmembership{Fellow, IEEE}, and Raymond Kai-yu Tong\textsuperscript{\dag}, \IEEEmembership{Senior Member, IEEE}
\thanks{This research was partly supported by Research Impact Fund (R5039-23F) from Research Grants Council of Hong Kong. \textsl{\textsuperscript{\dag}Corresponding authors: Zhe Xu and Raymond Kai-yu Tong.}}
\thanks{*: X. Wu and Z. Xu contributed equally to this work.}
\thanks{X. Wu, Z. Xu and R. Tong are with Department of Biomedical Engineering, The Chinese University of Hong Kong, Hong Kong, China.}
\thanks{Z. Xu is also with Department of Radiation Oncology, Columbia University Irving Medical Center and Data Science Institute, Columbia University, New York, NY, USA.}
\thanks{C. Chen is with Department of Electrical and Electronic Engineering and School of Biomedical Engineering, The University of Hong Kong, Hong Kong, China.}
\thanks{J. Ma is with Department of Computer Science, City University of Hong Kong, Hong Kong, China.}
\thanks{Y. Zheng is with Department of Artificial Intelligence, Westlake University, Hangzhou, China.}}


\markboth{Bidirectional Conservative-Radical Complementary Learning}%
{Wu \MakeLowercase{\textit{et al.}}: Bidirectional Conservative-Radical Complementary Learning}



\IEEEtitleabstractindextext{%
\begin{abstract}
Class-incremental learning (CIL) in medical image-guided diagnosis requires models to retain diagnostic expertise on previously learned disease categories while continually adapting to newly emerging ones, which is a key step toward scalable clinical deployment. This problem is particularly challenging due to heterogeneous clinical data and privacy constraints that preclude memory replay. Although pretrained foundation models (PFMs) have revolutionized general-domain CIL through transferable and expressive representations, their potential in medical imaging remains underexplored, where domain-specific adaptation is essential yet challenging due to anatomical complexity and inter-institutional heterogeneity. To bridge this gap, we first conduct a systematic benchmark of recent PFM-based CIL methods in the medical domain and further propose Bidirectional Conservative-Radical Complementary Learning (Bi-CRCL), a dual-learner framework inspired by the brain’s complementary learning systems. Bi-CRCL comprises two synergistic PFM-based learners: (i) a \textbf{conservative learner} (neocortex-like) that preserves accumulated diagnostic knowledge through stability-oriented updates, and (ii) a \textbf{radical learner} (hippocampus-like) that rapidly acquires new categories via plasticity-oriented adaptation. Specifically, the dual-learner cross-classification alignment mechanism harmonizes their complementary strengths, reconciling inter-task decision boundaries to mitigate catastrophic forgetting. At the core of Bi-CRCL lies a bidirectional design mirroring hippocampus–neocortex interaction: prior to each new task, the radical learner is initialized with the conservative learner’s consolidated weights (forward transfer); after adaptation, the radical learner’s updates are progressively integrated back into the conservative learner via exponential moving average (backward consolidation). This cyclic exchange enables the continual integration of new knowledge while preserving prior expertise. During task-agnostic inference, Bi-CRCL adaptively fuses outputs from both learners to achieve robust final predictions. Comprehensive experiments on five medical imaging datasets validate Bi-CRCL’s effectiveness over state-of-the-art methods. Further evaluations across different PFMs, severe cross-dataset distribution shifts, varying task granularities, and reversed task orders confirm its robustness, scalability, and strong generalization capacity. 
% The implementation is available at \url{https://github.com/CUHK-BMEAI/CRCL}.
\end{abstract}

}

\maketitle

% is not selected <OR> if conference mode is selected - because compsoc
% conference papers position the abstract like regular (non-compsoc)
\IEEEdisplaynontitleabstractindextext

% \fi
\IEEEpeerreviewmaketitle

\ifCLASSOPTIONcompsoc
\IEEEraisesectionheading{
\section{Introduction}
\label{sec:introduction}}
\else
%\section{Introduction}
%\label{sec:introduction}
\fi


% Plan:
% 1012: Intro, Related work
% 1014: Experiments
% 1014-1016: Methods

In medical image-guided disease diagnosis, it is essential to continually update diagnostic models to adapt to evolving healthcare data, especially as new disease categories emerge \cite{wu2024survey,AdapterCL2023}. This, in turn, requires models to incrementally integrate new clinical knowledge while preserving performance on previously learned conditions. However, conventional deep learning paradigms struggle with this requirement, as they are prone to catastrophic forgetting, the tendency to overwrite prior knowledge when trained on new tasks sequentially, which significantly hinders their scalability and long-term effectiveness. Ideally, diagnostic models should maintain high performance on both previously seen and newly introduced disease categories without access to explicit task identity during inference. This clinical need presents a significant challenge, commonly framed as class-incremental learning (CIL), that requires balancing the trade-off between stability (retaining existing diagnostic expertise) and plasticity (integrating novel disease patterns).

\begin{figure}[t]
  \centering
  \centerline{\includegraphics[width=0.48\textwidth]{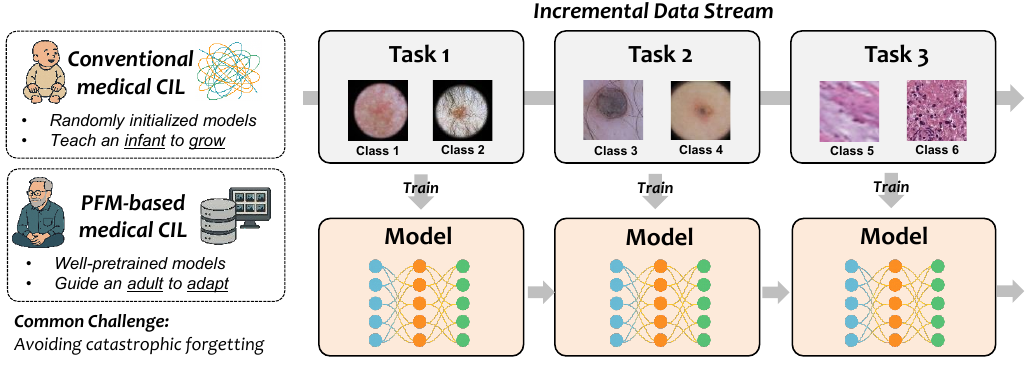}}
  \caption{Illustration of conventional vs. PFM-based medical CIL. Conventional CIL trains models from scratch, similar to teaching an infant to learn from zero experience. PFM-based CIL builds on PFMs, analogous to guiding an experienced adult to adapt efficiently to new tasks. Both paradigms share the fundamental challenge of catastrophic forgetting when learning new classes sequentially.}
  \label{Fig_comparison}
  \vspace{-0.3cm}
\end{figure}

Conventional CIL methods are generally grouped into three categories: replay-based methods, which store or synthesize previous samples to retrain on mixed data \cite{castro2018replay1,hou2019replay2,rebuffi2017icarl}; regularization-based methods, which penalize parameter updates that interfere with prior knowledge \cite{li2017reg1,kirkpatrick2017overcoming_reg3_EWC,wang2022foster}; and adaptive-architecture methods, which expand or selectively activate model components to accommodate new tasks \cite{abati2020conditional_arc1,yoon2017lifelong_arc2}. However, these methods typically assume models are trained from scratch and require extensive parameter optimization, akin to teaching an infant to grow into a radiologist (as illustrated in Fig. \ref{Fig_comparison}). This training-from-scratch paradigm hinders scalability and practical deployment, particularly in medical settings where data collection and computation are costly. More fundamentally, such methods struggle to maintain a unified and discriminative representation space across evolving disease categories, often leading to degraded generalization in long-term diagnostic scenarios.

Recent progress in pretrained foundation models (PFMs) has revitalized the CIL landscape by offering powerful and transferable representations learned from large-scale data \cite{L2P2022,zhou2024adam,tan2024shift,gao2023LAE,zhao2024safe}. Leveraging these generalized features allows continual learners to adapt efficiently without full retraining, resembling the process of guiding an experienced adult to specialize as a radiologist (Fig. \ref{Fig_comparison}). This paradigm shift enables faster adaptation, stronger representation reuse, and reduced catastrophic forgetting.
Among the approaches, prompt-based methods \cite{L2P2022,DualP2022,CODAP2023} introduce learnable context tokens to reuse frozen PFMs, yet managing growing prompt pools and ensuring compatibility with unseen classes increase complexity and make them brittle under domain shifts. Subsequent works explored broader adaptation strategies. SLCA \cite{zhang2023slca} introduces dual learning rates for backbone and classifier tuning, alongside a Gaussian-based classifier rectification strategy. LAE \cite{gao2023LAE} extends SLCA’s learning rate calibration by incorporating model merging to consolidate knowledge from earlier tasks. ADAM \cite{zhou2024adam} demonstrates the effectiveness of a prototypical classifier as a strong baseline, merging embeddings from a frozen PFM and a first-session adapted downstream model for subsequent classification. EASE \cite{zhou2024ease} and MOS \cite{sun2024mos} enhance feature representations by merging outputs from multiple task-specific adapters. Meanwhile, SSIAT \cite{tan2024shift} tackles the issue of feature drift by estimating class prototype shifts across tasks and enforcing a unimodal distribution assumption in replay-based unified training. 
Despite recent progress in natural image domains, PFM-based CIL remains largely underexplored in medical imaging, where the scarcity of medical-specific PFMs poses an additional bottleneck. In practice, most existing medical CIL studies continue to rely on non-PFM schemes, typically adopting replay-based retraining or full-model regularization. However, these paradigms face two critical limitations: (1) privacy risks, as storing and reusing patient data in memory buffers violates clinical data governance principles, and (2) computational inefficiency, as repeatedly applying parameter- and feature-level regularization to the entire model for every new task is prohibitively expensive. To bridge this gap, we harness general-domain PFMs as strong transferable backbones to take the initiative to benchmark recent PFM-based CIL methods on medical imaging datasets and, more importantly, to advance replay-free CIL strategies for evolving disease diagnosis. Although primarily grounded in general-domain PFMs for their superior generalizability, our framework remains compatible with medical-specific PFMs, offering a unified and extensible solution across diverse domains.

\begin{figure*}[t]
  \centering
  \centerline{\includegraphics[width=\textwidth]{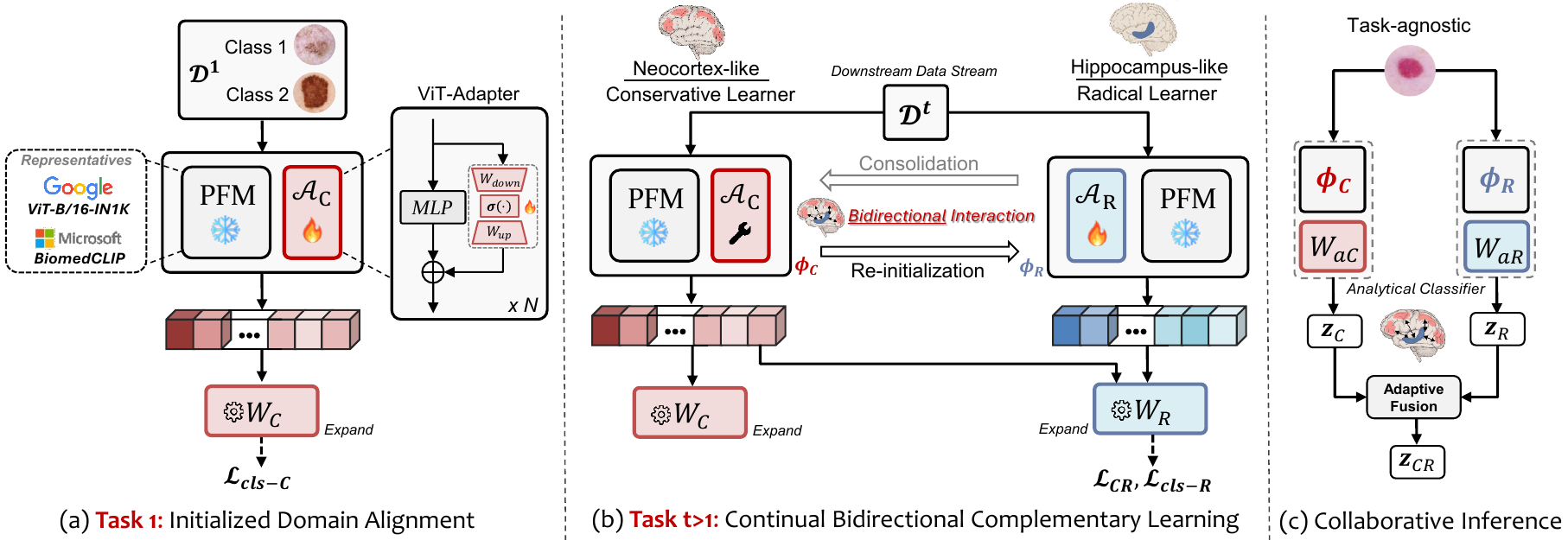}}
  \caption{Overview of our Bi-CRCL framework. (a) Initialized Domain Alignment: At task $t=1$, the general-domain PFM adapts to the medical domain via adapter tuning to enable efficient PFM transfer while preserving generalization; (b) Continual Bidirectional Complementary Learning: For $t>1$, a radical learner incrementally learns new classes and exchanges knowledge with the conservative learner through bidirectional consolidation and re-initialization, emulating hippocampus–neocortex coordination; (c) Collaborative Inference: During inference, outputs from both learners are projected via analytical classifiers and adaptively fused to produce robust task-agnostic predictions.}
  \label{fig_framework}
  % \vspace{-0.3cm}
\end{figure*}

To address the stability-plasticity dilemma and the intrinsic challenges of medical CIL, such as low inter- and intra-class variability, high inter-domain heterogeneity and strict privacy constraints, we propose Bidirectional Conservative-Radical Complementary Learning (Bi-CRCL), a brain-inspired framework grounded in the complementary learning systems (CLS) theory of human cognition \cite{kumaran2016cls} (as depicted in Fig. \ref{fig_framework}). Bi-CRCL comprises two synergistic and specialized learners built upon PFMs: (i) a conservative learner, analogous to the neocortex, which preserves accumulated diagnostic knowledge through stability-oriented parameter updates; and (ii) a radical learner, akin to the hippocampus, which rapidly acquires new disease concepts through plasticity-driven adaptation. 
To mitigate catastrophic forgetting while maintaining adaptability, Bi-CRCL introduces a dual-learner cross-classification alignment that harmonizes representations across tasks and reconciles evolving decision boundaries.
Central to our design is a bidirectional knowledge interaction between the two learners, inspired by the reciprocal pathways linking the hippocampus and neocortex. In the forward phase, before learning each new task, the radical learner is initialized with the conservative learner’s consolidated adapters, ensuring stable yet adaptable foundations. In the backward phase, after learning, the newly acquired representations are assimilated back into the conservative learner, consolidating long-term memory. This coordination enables continual adaptation without replay while preserving representational stability.
During inference, Bi-CRCL employs a closed-form analytical classifier to enhance class separability and adaptively fuses the complementary predictions of both learners for robust and task-agnostic diagnosis.
Extensive experiments on five medical imaging benchmarks demonstrate Bi-CRCL’s consistent superiority over state-of-the-art (SOTA) conventional and PFM-based CIL methods, with additional studies across diverse PFMs, severe cross-domain shifts, varying task granularities, and reversed task orders confirming its robustness, scalability, and generalization capability.

Our main contributions are as follows:
\begin{itemize}
    \item We benchmark recent PFM-based general-domain CIL methods on five medical imaging datasets. The evaluation reveals that PFMs substantially empower continual learning in medical diagnosis, even when pretrained on general-domain data, yet a notable gap remains due to the unique challenges of medical imaging such as low inter- and intra-class variability and high inter-domain heterogeneity. This benchmark provides a standardized foundation for advancing future research in medical CIL.
    \item We propose Bi-CRCL, a brain-inspired replay-free continual learning framework for medical image-guided diagnosis. It couples a radical and a conservative learner through bidirectional knowledge interaction and dual cross-classification alignment, balancing plasticity and stability. During inference, closed-form analytical classifiers and adaptive prediction fusion jointly enable robust and task-agnostic diagnostic performance.
    \item We conduct extensive experiments on five diverse medical imaging datasets, evaluating Bi-CRCL under standard class-incremental settings, varying task granularities, cross-dataset shifts and reversed task orders. Bi-CRCL consistently outperforms SOTA methods and remains competitive even without memory replay, demonstrating strong practicality in privacy-sensitive clinical scenarios. Moreover, our analysis shows that general-domain PFMs exhibit greater adaptability and transferability than medical-specific ones, highlighting their potential as a resilient foundation for continual medical diagnosis.
\end{itemize}

This work substantially extends our preliminary study \cite{wu2025crcl} at MICCAI'25 in three major aspects: (i) we further introduce a bidirectional knowledge interaction mechanism that more closely emulates the reciprocal pathways between the hippocampus and neocortex \cite{kumaran2016cls}, notably improving the coordination and robustness between the conservative and radical learners compared with the earlier single-directional design, which only consolidates knowledge from the radical to the conservative learner;
(ii) we enhance the inference stage with a confidence-based adaptive fusion strategy that more effectively integrates complementary predictions from the two learners for reliable diagnosis;
(iii) we conduct extensive analyses across more datasets, examining the impact of general- versus medical-domain PFMs, cross-dataset and cross-domain learning, task granularity, replay necessity, and task order. These comprehensive studies provide valuable insights and further confirm the robustness, scalability, and generalization capability of our method.

\textbf{The paper is organized as follows:} 
Sec.~\ref{sec_2} reviews related work on class-incremental learning in both general- and medical-domain contexts, with an emphasis on recent PFM-based approaches.
Sec.~\ref{sec_3} introduces our Bi-CRCL framework, comprising three core components: (i) adapter-based first-session domain alignment, (ii) continual bidirectional complementary learning across sessions, and (iii) collaborative inference through analytical classifiers and adaptive prediction fusion.
Sec.~\ref{sec_4} presents extensive experiments on five medical imaging benchmarks across diverse classification tasks and modalities, and provides in-depth analyses under cross-domain and inter-dataset settings, varying task granularities and task orders. It also compares general- and medical-domain PFMs and includes replay and ablation studies on key framework components.
Finally, Sec.~\ref{sec_discuss} discusses the clinical relevance, limitations, and future directions of PFM-based continual learning in medical diagnosis.

%\IEEEpubidadjcol
\section{Related Works}
\label{sec_2}
\subsection{Class-incremental Learning}
Conventional CIL methods can be broadly categorized into three main paradigms \cite{wu2024survey}:
(i) Replay-based approaches, which preserve a subset of past samples or regenerate pseudo-data to mitigate forgetting~\cite{castro2018replay1,hou2019replay2,rebuffi2017icarl};
(ii) Regularization-based methods, which constrain parameter updates to maintain previously acquired knowledge~\cite{li2017reg1,kirkpatrick2017overcoming_reg3_EWC,wang2022foster}; and
(iii) Dynamic architecture strategies, which expand or modulate network capacity to accommodate new tasks~\cite{abati2020conditional_arc1,yoon2017lifelong_arc2}.
While effective to some extent, these methods typically require training models from scratch and extensive hyperparameter tuning, which limits their scalability and practical deployment.
Conceptually, such approaches resemble teaching an infant to grow and learn new knowledge from the beginning, where the model must relearn fundamental representations for each task.
In contrast, the emergence of pretrained foundation models (PFMs) has reshaped the CIL paradigm. PFMs provide strong and generalizable representations that enable rapid adaptation to novel tasks without full retraining~\cite{L2P2022,DualP2022,CODAP2023,zhou2024adam,tan2024shift,gao2023LAE,zhou2024ease,sun2024mos,wu2025crcl,wurandom}.
From this perspective, PFM-based CIL is akin to guiding an experienced adult to acquire new skills, emphasizing efficient adaptation rather than foundational learning.
Early PFM-based CIL methods such as L2P \cite{L2P2022}, DualPrompt \cite{DualP2022}, and CODAPrompt \cite{CODAP2023} introduce learnable prompts to steer frozen PFMs toward new tasks, but managing prompt pools and unseen-class compatibility remains challenging. Subsequent studies like SLCA \cite{zhang2023slca}, LAE \cite{gao2023LAE}, ADAM \cite{zhou2024adam}, and EASE \cite{zhou2024ease} explored calibrated tuning, model merging, and adapter fusion to improve representation reuse and mitigate forgetting. MOS \cite{sun2024mos} and SSIAT \cite{tan2024shift} further enhance stability via multi-adapter ensembles and prototype-shift rectification. Overall, these works mark a transition from memory-intensive retraining to parameter-efficient continual adaptation, where the PFM acts as a stable representational anchor.
Despite remarkable success in natural image domains, these methods often perform suboptimally in medical image-guided diagnosis, where limited data diversity, domain heterogeneity, and subtle inter-class variations intensify representation drift and forgetting.

\subsection{Class-incremental Learning in Medical Imaging}
Medical CIL has traditionally relied on training models from scratch, typically through replay-based retraining or parameter-intensive regularization to mitigate catastrophic forgetting~\cite{Perkonigg2021,Byun2023,Bai2023,Chee2023,wu2023continual,LiDOCTOR2023,Yeganeh2023,Sadafi2023}.
With the advent of PFMs offering strong generalization, recent general-domain studies have shifted toward balancing stability and plasticity during continual adaptation.
However, PFM-based CIL remains largely underexplored in medical imaging. The Adapter-based Continual Learning (ACL) framework~\cite{AdapterCL2023} marks an early effort to insert lightweight adapters into a fixed ResNet-18 backbone for incremental disease classification. While this design reduces full-model retraining cost, it remains replay-based, requiring stored patient images for continual updates, and its shallow CNN backbone limits scalability to transformer-based PFMs widely used in current practice.
Meanwhile, the scarcity of domain-specific PFMs poses an additional bottleneck. Although medical-specific PFMs such as BiomedCLIP~\cite{zhang2023biomedclip} and RAD-DINO~\cite{rad-dino2025} have been proposed, they often exhibit limited transferability across sub-domains. Empirical comparisons further indicate that domain specialization does not guarantee stronger medical representations (Table~\ref{tab:ours_medptm}), as medical PFMs frequently yield less discriminative features than general-domain counterparts like ViT-B/16-IN1K~\cite{alexey2020image}.
Given their superior generalization and representational stability, we primarily adopt general-domain PFMs as the backbone for our study. Building on this, we propose the brain-inspired Bi-CRCL, which bridges general-domain PFMs with clinical demands and enables replay-free, robust continual adaptation to evolving disease diversity.

\section{Methodology}
\label{sec_3}
\subsection{Preliminary}
\textbf{Problem Definition.} CIL is defined as training a model on a sequential data stream where new classes are introduced incrementally over time. Formally, the training set for task \textit{t} is denoted as 
$\mathcal{D}^t=\left\{\left(x_{i,t}, y_{i,t}\right)\right\}_{i=1}^{n_t}$, where $t \in\{1,2, \ldots, T\}$ represents the task index among a total of $T$ tasks, and each task contains $n_t$ instances. Here, $x_{i,t}$ denotes the $i$-th input sample of task $t$, and $y_{i,t}$ is its corresponding ground-truth label. We assume that each task $t$ introduces a unique set of classes $Y_t$, with no overlap between tasks: $Y_t \cap Y_{t^{\prime}}=\varnothing$ for $t \neq t^{\prime}$. The goal is for the model to perform well on a test set encompassing all classes introduced up to task $t$, i.e., $\mathcal{Y}_t=Y_1 \cup \cdots Y_t$.

\noindent\textbf{PFM-based CIL.} We aim to build a model $f(\mathbf{x}): X \rightarrow \mathcal{Y}_t$ that can learn new classes incrementally without forgetting previously learned ones. In line with prior replay-free PFM-based CIL studies \cite{tan2024shift,zhang2023slca,zhou2024adam,zhou2024ease}, we consider a pre-trained Vision Transformer (ViT) foundation model is available for initializing $f(\mathbf{x})$. This model is then decomposed into a feature extraction backbone $\mathcal{F}_{\theta_{\mathrm{bne}}}$ and a linear classification layer $f_{\theta_{\mathrm{cls}}}$. The backbone $\mathcal{F}_{\theta_{\mathrm{bne}}}$ serves as a feature encoder $\phi(\cdot):\mathbb{R}^D \rightarrow \mathbb{R}^d$, mapping input images to feature embeddings, while the classifier layer $f_{\theta_{\mathrm{cls}}}$, parameterized by a weight matrix $W \in \mathbb{R}^{d \times\left|Y_t\right|}$, maps the feature embeddings to class logits. The model can then be formulated as $f(\mathbf{x})=W^{\top} \phi(\mathbf{x})$, where $W=\left[\mathbf{w}_1, \mathbf{w}_2, \cdots, \mathbf{w}_j
\right]$ and $\mathbf{w}_j$ denotes the classifier weight vector for class $j$. For ViT-based backbones, we use the embedded \texttt{[CLS]} tokens as the feature embeddings $\phi(\mathbf{x})$.

\noindent\textbf{Tuning with Adapter.} Recent studies have explored a range of parameter-efficient tuning (PET) methods to adapt PFMs for downstream tasks, including SSF \cite{Lian_2022_SSF}, VPT \cite{jia2022vpt}, and adapters \cite{adapterPET}. However, our empirical findings indicate that SSF and VPT tend to suffer from overfitting to the current distribution, resulting in unstable performance on complex medical datasets. Based on this observation, we advocate adapters as the PET strategy in our PFM-based CIL framework. Specifically, adapters are lightweight modules composed of a down-projection $W_{\text {down}} \in \mathbb{R}^{k \times \hat{k}}$, a ReLU activation function $\sigma$, and an up-projection $W_{u p} \in \mathbb{R}^{\hat{k} \times k}$, forming a bottleneck structure. Following \cite{tan2024shift,zhou2024ease}, we integrate these adapters into the multilayer perceptron (MLP) layers of the ViT, as depicted in Fig. \ref{fig_framework}(a), using a projected dimension $\hat{k}=64$ \cite{adapterPET}. Let $x_{in}$ denote the input of the MLP layer, the output of the adapter-augmented MLP is:
$x_{out}=MLP(x_{in})+\sigma\left(x_{in}W_{\text {down }}\right)W_{u p}$,
where matrix multiplication is implied. During training, the PFM backbone remains frozen, and only the adapters and the classification heads are updated. Hence, the optimizable parameter set can be denoted as: $\Theta=\theta_{W_{\text {down}}} \cup \theta_{W_{\mathrm{up}}} \cup \theta_W$, where $\theta_W$ corresponds to the parameters of the classification heads. This adapter-based ViT fine-tuning forms the foundation of our replay-free PFM-based CIL framework, enabling flexible and parameter-efficient adaptation to downstream medical tasks while mitigating overfitting due to the relatively small size of each incremental dataset.

\subsection{Bidirectional Conservative-Radical Complementary Learning}
\noindent\textbf{Framework Overview.}
Balancing plasticity and stability is a fundamental challenge in PFM-based CIL, and this challenge is further exacerbated when transferring generalized representations from general-domain PFMs to the medical domain. To address this, our Bi-CRCL framework (Fig. \ref{fig_framework}) consists of three key processes: (i) initialized domain alignment at the first task, which adapts the PFM to the medical domain via direct adapter tuning, enabling efficient PFM transfer while preserving generalization; (ii) continual bidirectional complementary learning for subsequent tasks ($t>1$), inspired by the two-way hippocampus-neocortex interaction in the human brain, where a radical learner incrementally learns new classes while exchanging knowledge with the conservative learner through bidirectional consolidation and re-initialization, effectively balancing knowledge retention and adaptation; and (iii) collaborative inference, where outputs from both learners are projected via analytical classifiers and adaptively fused to produce robust and task-agnostic predictions.

\noindent\textbf{Initialized Domain Alignment.}
As illustrated in Fig. \ref{fig_framework}(a), we perform initialized domain alignment in the first incremental task ($t=1$) to bridge the gap between the utilized PFM (ImageNet-21K\&1K-pretrained ViT-B/16-IN1K by default) and downstream medical datasets. This is achieved by adapting the PFM to the medical domain via lightweight ViT-Adapter tuning. Specifically, the PFM backbone is frozen to preserve generalization, while only the adapters $\mathcal{A}_{C}$ and the classifier $W_C$ are updated. Notably, unlike a conventional trainable classification layer, $W_C$ expands to incorporate new classes and is updated using class prototypes as imprinted weights \cite{qi2018imprinted}. The \textit{gear} icon in Fig. \ref{fig_framework} visually represents this distinction, indicating that $W_C$ undergoes structured updates rather than standard parameter learning. In the first task, no additional parameter constraints are imposed to ensure sufficient domain adaptation, as the performance of the PFM on its pretraining task is not our primary concern. Note that this step is optional and can be adjusted if the pretraining task of the PFM is itself a focus of interest. Our training objective is to minimize the cross-entropy (CE) loss: 
\begin{equation}
    \label{eq:clsloss_C}
    \mathcal{L}_{\text {cls-C }}=\frac{1}{N_b} \sum_{i=1}^{N_b} \mathrm{CE}\left(W_{C}^{\top} \phi_{C}\left(x_i\right), y_i\right),
\end{equation}
where $N_b$ is the batch size, $\phi_C(x_i)$ denotes the adapted feature embeddings for input $x_i$, and $y_i$ is the corresponding ground-truth label.

\noindent\textbf{Continual Bidirectional Complementary Learning.}
Fig. \ref{fig_framework}(b) illustrates the continual bidirectional complementary learning paradigm (for $t>1$) that balances stability and plasticity through the interaction of radical and conservative learners. The radical learner, akin to the hippocampus, rapidly encodes new information by optimizing all parameters of its dedicated adapters, $\mathcal{A}_{R}$, via backpropagation in each incremental training task. Meanwhile, the conservative learner, reminiscent of the neocortex, preserves consolidated knowledge while gradually integrating new patterns. In the human brain, bidirectional pathways support the coordination between the hippocampus and neocortex to enable fast learning and long-term consolidation of knowledge \cite{kumaran2016cls}. Inspired by this, in our framework, we introduce a bidirectional knowledge interaction mechanism between the radical and conservative learners. In the forward direction, at the start of each new task, the radical learner’s adapters $\mathcal{A}_{R}$ are initialized using the continuously consolidated adapters of the conservative learner, i.e., $\mathcal{A}_{C}$. This initialization provides a warm start for learning new information, allowing the radical learner to build upon prior long-term knowledge. In the backward direction, after learning, the updated $\mathcal{A}_{R}$ feeds new knowledge back into the $\mathcal{A}_{C}$. Specifically, the conservative learner's adapters $\mathcal{A}_{C}$ are updated via an exponential moving average (EMA) of the radical learner’s adapters (represented by a \textit{wrench} icon in Fig. \ref{fig_framework}), ensuring stable and incremental assimilation. Formally, in the $t$-th ($t>1$) task, this consolidation is defined as: 
\begin{equation}
    \theta_{\mathcal{A}_{C,t}} = \alpha \theta_{\mathcal{A}_{C,{t-1}}} + (1-\alpha) \theta_{\mathcal{A}_{R,t}},
\end{equation}
where $\alpha$ is the EMA decay rate and empirically set to 0.99 \cite{xu2023ambiguity}. Together, this bidirectional interaction forms a stable and biologically inspired learning loop, where the conservative learner offers long-term guidance, and the radical learner continuously contributes task-specific information, striking an effective balance between plasticity and stability. 

To learn new tasks, the radical learner is primarily optimized using a standard cross-entropy classification loss:
\begin{equation}
    \mathcal{L}_{\text {cls-R}}=\frac{1}{N_b} \sum_{i=1}^{N_b} \mathrm{CE}\left(W_R^{\top} \phi_R(x_i), y_i\right),
\end{equation}
where $W_R$ represents the structurally updated classification weights (analogous to $W_C$), and $\phi_R(x_i)$ denotes the radical learner’s feature embeddings for input $x_i$. At task $t$, $W_R \in \mathbb{R}^{d \times\left|Y_{t-1}\right|}$ expands to $W_R \in \mathbb{R}^{d \times\left|Y_{t}\right|}$ with the class prototypes of newly introduced classes. To promote compatibility between the learners' representations and facilitate feature alignment, we employ the cross-classification regularization loss $\mathcal{L}_{CR}$. This loss encourages the conservative learner’s features to remain compatible with the radical classifier, reinforcing consistency in representation learning of the radical learner and mitigating severe semantic shifts that could lead to biased decision boundaries:
\begin{equation}
\mathcal{L}_{CR}=\frac{1}{N_b} \sum_{i=1}^{N_b} \mathrm{CE}\left(W_R^{\top}\phi_C(x_i), y_i\right).
\end{equation} 
Overall, the final loss for the radical learner is: 
\begin{equation}
\mathcal{L}_{R}=\mathcal{L}_{\text{cls-R }}+\mathcal{L}_{CR}.
\end{equation} 
This combined objective enables the radical learner to effectively acquire new knowledge while maintaining alignment with the conservative learner, fostering a stable process.

\begin{algorithm}[t]
\caption{Bi-CRCL}
\label{alg:bicrcl}
\begin{algorithmic}[1]
\REQUIRE Tasks $\{\mathcal{D}^t\}_{t=1}^{T}$, PFM (frozen backbone), EMA decay $\alpha$
\ENSURE Conservative learner adapter $\mathcal{A}_C$, radical learner adapter $\mathcal{A}_R$

\STATE \textbf{Domain alignment ($t{=}1$):} Train $\mathcal{A}_C$ on $\mathcal{D}^1$ with $\mathcal{L}_{\text{cls-C}}$ 

%\STATE \textbf{Inference ($t{=}1$):} Obtain logits $z_C$ and final prediction $y^*$

\FOR{$t = 2$ to $T$}
    \STATE \textbf{Forward transfer (init $\,\mathcal{A}_R$):} $\mathcal{A}_R \gets \mathrm{copy}(\mathcal{A}_C)$
    \STATE \textbf{Radical update:} Train $\mathcal{A}_R$ on $\mathcal{D}^t$ using loss $\mathcal{L}_{R}=\mathcal{L}_{\text{cls-R}}+\mathcal{L}_{CR}$

    \STATE \textbf{Backward consolidation:} 
    \FOR{each parameter $\theta$ in $\mathcal{A}_C$}
        \STATE $\theta_{\mathcal{A}_{C,t}} = \alpha \theta_{\mathcal{A}_{C,{t-1}}} + (1-\alpha) \theta_{\mathcal{A}_{R,t}}$
    \ENDFOR
    \STATE \textbf{Update analytical classifiers:} Update $W_{aC},W_{aR}$ via Eq.~(\ref{eq:analyticalweights})
    
\ENDFOR
\STATE \textbf{Collaborative inference (any $t$):} Obtain logits $z_C$ and $z_R$; Adaptively fuse $z_{CR}$ to obtain final prediction $y^*$.

\end{algorithmic}
\end{algorithm}

\noindent\textbf{Analytical Classifier Learning.}
A common choice in CIL is the class-prototype classifier \cite{zhou2024adam}, which computes class means and applies the nearest-class-mean (NCM) rule. While effective in standard settings, this design assumes that class embeddings are isotropic and well separated. In medical CIL, however, prototypes often exhibit strong inter-class correlations due to domain shifts and limited inter- and intra-class variability, leading to representation collapse and degraded discriminability. To alleviate these issues, we adopt the analytical learning paradigm \cite{zhuang2022acil}, which formulates classifier construction as a ridge regression problem \cite{hoerl1970ridge}, yielding a closed-form solution that can be updated recursively without re-accessing past data, enabling efficient expansion to new tasks while theoretically being guaranteed to yield a classifier analytically equivalent to retraining from scratch on all previously seen data. This paradigm consists of two stages:
(1) a randomized feature transformation that introduces minimal perturbation to the embedding space while improving class separability, and
(2) a closed-form classifier estimation that derives weight parameters analytically from accumulated feature statistics.
Unlike gradient-based updates, this analytical formulation avoids iterative optimization and thereby mitigates issues such as catastrophic forgetting and gradient instability. Analytical classifiers have also shown promising performance in few-shot and imbalanced continual learning settings \cite{zhuang2023gkeal, fang2024air}, motivating their use in data-scarce medical domains.

Specifically, for each learner $l$, we extract embeddings $\phi_{l}(\mathbf{x}) \in \mathbb{R}^d$ from the frozen adapter-finetuned backbone and map them into a higher-dimensional feature space using a fixed random projection matrix $W_{\text{rand}} \in \mathbb{R}^{d \times M}$ sampled from $\mathcal{N}(0, 1)$, followed by a ReLU activation function:
\begin{equation}
h_l = \mathrm{ReLU}\left(\phi_{l}(\mathbf{x})^{\top} W_{\text{rand}}\right) \in \mathbb{R}^M \quad (M>d).
\end{equation}
We then align the transformed features with class labels through ridge regression, minimizing a regularized least-squares objective:
\begin{equation}
\arg\min_{W_l} \left\| Y - H_{l} W_l \right\|_F^2 + \beta \left\| W_l \right\|_F^2, 
\end{equation}
where \( H_{l}\in \mathbb{R}^{N \times M} \) is the stacked matrix of all projected sample features; \( Y \in \mathbb{R}^{N \times |Y_t|} \) contains the corresponding one-hot labels; $N = \sum_{i=1}^{t} N_i$ is the total number of samples accumulated up to the current task; \( \|\cdot\|_F \) is the Frobenius norm; and \( \beta \) is the ridge regularization parameter selected by cross-validation-based optimization. The closed-form solution, obtained through recursive least-squares estimation using accumulated feature statistics across sessions 1 to $t$, is given by:
\begin{equation}
\label{eq:analyticalweights}
\hat{W}_l = \left( H_{l}^\top H_{l} + \beta I \right)^{-1} H_{l}^\top Y \in \mathbb{R}^{M \times |Y_t|},
\end{equation}
where $I$ is the identity matrix. The inverse term $(H_{l}^\top H_{l} + \beta I)^{-1}$ reweights the eigendirections of the feature space, functioning analogously to whitening transformation: it suppresses dominant shared components (the primary source of inter-class correlation) while amplifying task-specific variations. As a result, the analytical classifier yields decorrelated and more discriminative prototypes, enhancing robustness to distribution shifts commonly encountered in medical continual learning. The resulting classifier weights $\hat{W}_l$ are denoted as $W_{aC}$ and $W_{aR}$ for the conservative and radical learners, respectively. The logits for the projected features $h_l$ are then computed as: $z_l=h_l \hat{W}_{l}\in \mathbb{R}^{\left|Y_t\right|}$.

\noindent\textbf{Collaborative Inference.}
During inference, the conservative and radical learners output logits 
$z_C, z_R \in \mathbb{R}^{|Y_t|}$ for a given test image, 
where $|Y_t|$ denotes the cumulative number of classes observed so far. 
To assess their agreement, each logit is converted into a temperature-scaled probability distribution 
$\pi_C=\mathrm{softmax}(z_C/\tau)$ and $\pi_R=\mathrm{softmax}(z_R/\tau)$, where $\tau{=}0.1$ is used to sharpen the distributions, thereby emphasizing class agreement and penalizing mismatches. 
The symmetric KL divergence measuring the bidirectional disagreement between the two learners is computed as:
\begin{equation}
D_{\text{sym}}=\tfrac{1}{2}\!\left(D_{\mathrm{KL}}(\pi_C\parallel\pi_R)
+D_{\mathrm{KL}}(\pi_R\parallel\pi_C)\right).
\end{equation}
A dynamic threshold estimated from batch statistics distinguishes confident consensus from uncertain conflict:
\begin{equation}
\theta_{\text{div}}=\mathbb{E}[D_{\text{sym}}]+\lambda\,\mathrm{Std}(D_{\text{sym}}),
\end{equation}
where $\lambda{=}0.5$ by default. 
When the disagreement is small ($D_{\text{sym}}\le\theta_{\text{div}}$), we adopt the prediction from the more confident learner, determined by its maximum class probability. 
When the disagreement exceeds the threshold, both learners contribute via a confidence-weighted combination. These two cases can be unified as:
\begin{equation}
z_{CR}=(1-g)\,z_{m}+g\,(\alpha_C z_C+\alpha_R z_R),
\end{equation}
where $g=\mathds{1}[D_{\text{sym}}>\theta_{\text{div}}]$, $z_m$ denotes the logits from the more confident learner, and $\alpha_C$ and $\alpha_R$ are normalized confidence scores ($\alpha_C+\alpha_R=1$). 
The final prediction is then obtained as:
\begin{equation}
y^*=\arg\max(z_{CR}).
\end{equation}
This gating mechanism further balances stability and plasticity: 
when the two learners agree, the system follows its most reliable source of knowledge; 
when disagreement arises, it leverages bidirectional cooperation to refine uncertain predictions.

\section{Experiments}
\label{sec_4}

\begin{table*}[t]
\caption{Overview of medical image classification datasets.}
\label{table_data}
\centering
\scalebox{0.8}{
\begin{tabular}{p{3.5cm}|p{2cm}<{\centering}|p{2.3cm}<{\centering}|p{2cm}<{\centering}|p{2.5cm}<{\centering}|p{2cm}<{\centering}}
\Xhline{1pt} \textbf{Dataset} & \textbf{Classes} & \textbf{Training set} & \textbf{Test set} & \textbf{Task Num.} & \textbf{Size} \\\hline
Colon \cite{kather2019predicting} & 9 & 70,000 & 30,000 & 4 & $224 \times 224$ \\
Blood \cite{acevedo2019recognition} & 8 & 11,965 & 5,127 & 4 & $360 \times 363$ \\
Skin8 \cite{tschandl2018ham10000} & 8 & 3,555 & 705 & 4 & {$[600,1024]$} \\
MedMNIST-Sub \cite{medmnistv2} & 36 & 302,002 & 75,659 & 4 & $28 \times 28$ \\
COVID (CT\&X-ray) \cite{wang2023covid} & 11 & 3,939 & 1,072 & 6 & $224 \times 224$ \\
\Xhline{1pt}
\end{tabular}}
% \vspace{-0.02cm}
\end{table*}

\begin{figure}[t]
  \centering
  \centerline{\includegraphics[width=0.5\textwidth]{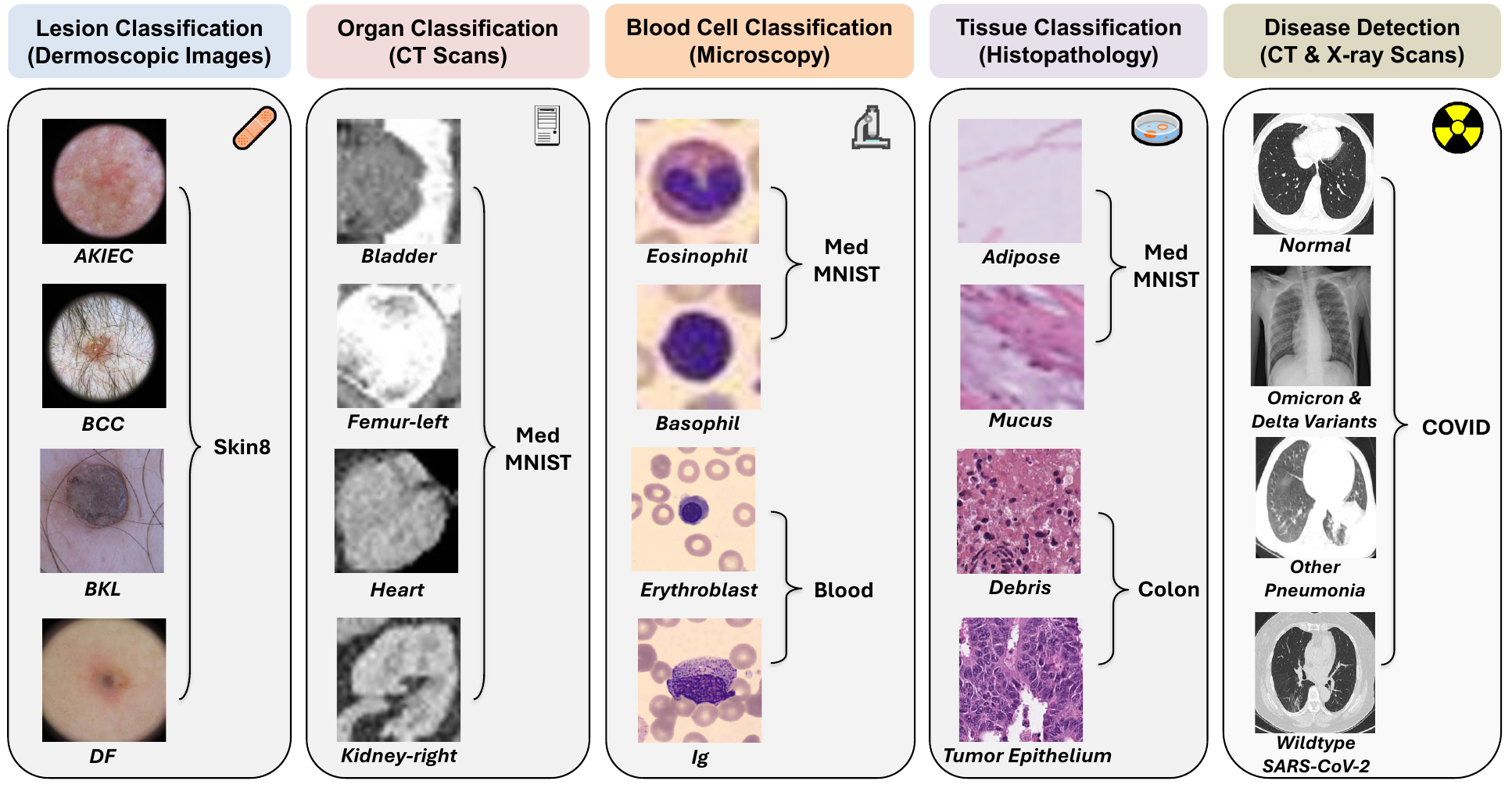}}
  \caption{An illustration of the diverse medical imaging tasks addressed by our model, spanning lesion classification (dermoscopic images), organ classification (CT scans), blood cell classification (microscopy), tissue classification (histopathology), and disease detection (CT \& X-ray scans), with representative example classes from each dataset.}
  \label{Fig_dataset_illustration}
  % \vspace{-0.3cm}
\end{figure}

\begin{table*}[t]
\caption{Performance on five medical datasets. ``$\dagger$" indicates the necessity of replaying prior data. The best and second-best results are \textbf{bolded} and \underline{underlined}, respectively.}
\label{tab_main}
\centering
\scalebox{0.85}{
\begin{tabular}{lp{1.4cm}<{\centering}p{1.4cm}<{\centering}|p{1.4cm}<{\centering}p{1.4cm}<{\centering}|p{1.4cm}<{\centering}p{1.4cm}<{\centering}|p{1.4cm}<{\centering}p{1.4cm}<{\centering}|p{1.4cm}<{\centering}p{1.4cm}<{\centering}}
\Xhline{1pt}
\multirow{2}{*}{\textbf{Method}} & \multicolumn{2}{c}{\textbf{Colon}} & \multicolumn{2}{c}{\textbf{Blood}} & \multicolumn{2}{c}{\textbf{Skin8}} & \multicolumn{2}{c}{\textbf{MedMNIST-Sub}} & \multicolumn{2}{c}{\textbf{COVID (CT\&X-ray)}} \\
\cmidrule(r){2-3} \cmidrule(r){4-5} \cmidrule(r){6-7} \cmidrule(r){8-9} \cmidrule(r){10-11} 
 & $Acc_{\text{Avg}}$ & $Acc_{\text{Last}}$ & $Acc_{\text{Avg}}$ & $Acc_{\text{Last}}$ & $Acc_{\text{Avg}}$ & $Acc_{\text{Last}}$ & $Acc_{\text{Avg}}$ & $Acc_{\text{Last}}$ & $Acc_{\text{Avg}}$ & $Acc_{\text{Last}}$ \\ 
\midrule
Joint Training &  - & 99.98 & - & 99.61 & - & 67.73 & - & 73.61 & - & 92.43 \\ 
Finetune & 38.65 & 10.43 & 37.47 & 14.54 & 39.62 & 17.87 & 29.18 & 5.66  & 26.55 & 10.54 \\ 
\midrule
FOSTER$^\dagger$ \cite{wang2022foster} & 90.81 & 86.69 & 87.70 & 90.24 & 55.00 & 39.01 & 58.24 & 31.46 & 71.65 & 55.50 \\
iCaRL$^\dagger$ \cite{rebuffi2017icarl} & 80.50 & 78.13 & 81.41 & 81.57 &  58.45 & 39.57 & 68.23 & 38.44 & 70.10 & 65.21 \\
DER$^\dagger$ \cite{yan2021DER} & 85.92 & 85.63 & 86.48 & 87.10 & 48.57 & 23.55 & 69.97 & 42.11 & 31.40 & 27.05 \\
\midrule
ACL$^\dagger$ \cite{AdapterCL2023} & 85.57 & 85.16 & 76.09 & 82.33 & 58.30 & 57.66 & 81.33 & 65.40 & 85.13 & 69.23 \\
L2P \cite{L2P2022} & 70.13 & 49.91 & 86.46 & 76.15 & 55.39 & 35.89 & 56.24 & 28.96 & 43.48 & 20.06 \\
DualPrompt \cite{DualP2022} & 79.47 & 63.36 & 76.62 & 66.27 & 52.32 & 27.38 & 54.92 & 26.31 & 43.62 & 19.68 \\
CODAPrompt \cite{CODAP2023} & 81.96 & 66.70 & 72.64 & 61.57 & 50.17 & 32.48 & 61.12 & 28.65 & 36.67 & 20.88 \\
LAE \cite{gao2023LAE} & 71.16 & 49.68 & 55.51 & 33.94 & 49.52 & 24.40 & 48.46 & 18.39 & 46.26 & 21.83 \\
SimpleCIL \cite{zhou2024adam} & 90.10 & 85.41 & 83.85 & 79.79 & 56.61 & 38.30 & 68.07 & 50.63 & 75.84 & 57.37 \\ 
ADAM-Adapter \cite{zhou2024adam} & 86.01 & 78.00 & 88.09 & 83.52 & 59.82 & 41.84 & 70.97 & 53.11 & 79.19 & 61.10 \\ 
SLCA \cite{zhang2023slca} & 86.66 & 76.73 & 90.23 & 82.29 & 58.91 & 40.71 & 56.39 & 44.42 & 63.05 & 60.82 \\ 
EASE \cite{zhou2024ease} & 89.62 & 82.48 & 68.85 & 67.60 & 60.40 & 40.43 & 65.11 & 39.26 & 78.47 & 59.98 \\ 
SSIAT \cite{tan2024shift} & 75.36 & 66.34 & 86.00 & 84.63 & 60.46 & 41.99 & 59.43 & 25.79 & 72.00 & 60.17  \\ 
MOS \cite{sun2024mos} & 91.46 & 87.60 & 92.53 & 90.18 & 68.54 & 51.77 & 74.59 & 51.80 & 89.96 & 80.60 \\  \hline
CRCL (ours) \cite{wu2025crcl} &  \underline{98.16} & \underline{97.58} &  \underline{97.13} & \underline{96.04}  & \underline{73.76} & \underline{61.32} & \textbf{84.70} & \underline{66.46} & \underline{95.49} & \underline{84.70} \\
Bi-CRCL (ours) &  \textbf{99.12} & \textbf{98.51} &  \textbf{98.08} & \textbf{97.56}  & \textbf{74.59} & \textbf{61.99} & \underline{84.22} & \textbf{69.71} & \textbf{96.12} & \textbf{88.15} \\
\Xhline{1pt}
\end{tabular}}
\vspace{-0.2cm}
\end{table*}
% M:
% Colon and Blood: 768
% Skin8: 2048
% MedMNIST: 2048? Now 1024
\begin{figure*}[t]
\centering
\includegraphics[width=1\textwidth]{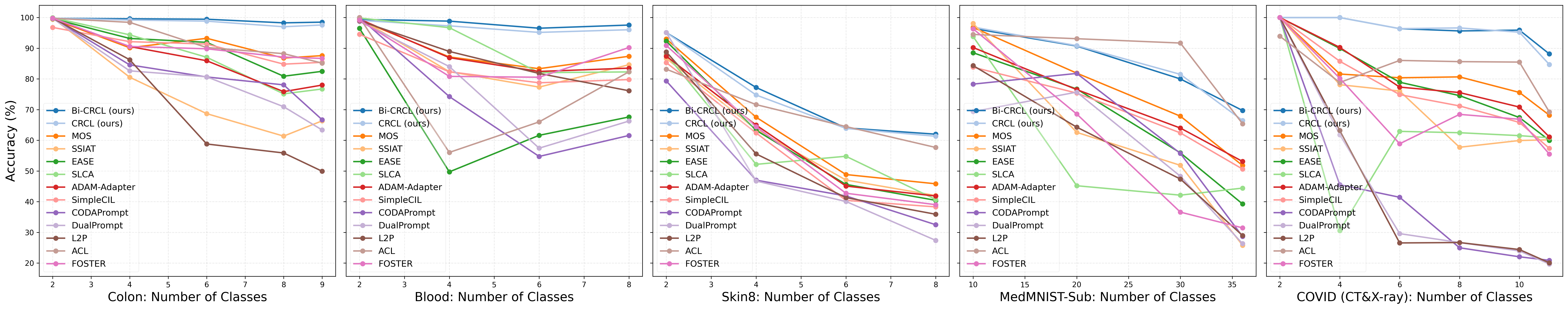}
\caption{The performance curves across learning sessions on five medical datasets. Bi-CRCL consistently achieves the best overall performance and exhibits minimal degradation as class number increases.}
\label{fig_results}
\end{figure*}
\subsection{Datasets and Experimental Setup}
\subsubsection{Datasets} 
We benchmark recent CIL methods on five diverse medical imaging datasets, summarized in Table~\ref{table_data}. The Colon dataset~\cite{kather2019predicting} comprises H\&E-stained histopathology images of colorectal cancer and healthy tissues, whereas the Blood dataset~\cite{acevedo2019recognition} contains peripheral blood cell images captured from blood smears. Skin8~\cite{tschandl2018ham10000}, derived from the ISIC challenge for dermatoscopic skin lesion classification, features substantial class imbalance. MedMNISTv2~\cite{medmnistv2} is a standardized biomedical benchmark containing 12 2D and 6 3D datasets for multi-class, multi-label, and ordinal regression tasks. Following previous work \cite{AdapterCL2023}, we use a subset of four 2D multi-class classification datasets—BloodMNIST, OrganAMNIST, PathMNIST, and TissueMNIST—collectively referred to as MedMNIST-Sub, adopting the same data splits as in earlier studies \cite{bayasi2024continual,AdapterCL2023}. The COVID (CT \& X-ray) dataset~\cite{wang2023covid} focuses on respiratory disease diagnosis using CT and X-ray scans. During training, all images are resized to $224 \times 224$ pixels. As illustrated in Fig.~\ref{Fig_dataset_illustration}, these datasets collectively cover a wide range of medical imaging tasks, including lesion, organ, blood cell, tissue, and disease classification, across multiple modalities such as microscopy, histopathology, dermoscopy, CT, and X-ray.

\subsubsection{Implementation and Evaluation Protocol} 
The framework is implemented on PyTorch using two NVIDIA A100 80G GPUs. Following \cite{zhou2024ease,zhou2024adam}, we adopt ViT-B/16-IN1K \cite{alexey2020image} as our default PFM, which is pre-trained on ImageNet-21K and fine-tuned on ImageNet-1K. General-domain PFMs remain the mainstream choice whereas medical-specific PFMs are still relatively underexplored. To examine the generalization capability of our framework, in Sec.~\ref{sec:medptm}, we further analyze the performance sensitivity and comparative advantages of using general-domain versus medical-specific PFMs. We conduct all experiments with a batch size of 48 for 20 epochs in the initial adaptation and 15 epochs in subsequent sessions, using stochastic gradient descent (SGD) with momentum and a cosine-annealed learning rate starting at 0.01. We apply random flipping and rotating for weak data augmentation. Following \cite{tan2024shift,zhang2023slca,zhou2024adam,zhou2024ease}, we report the last session accuracy $Acc_{\text {Last }}$ and the average accuracy across all incremental sessions, formulated as: $Acc_{\text{Avg}}=\frac{1}{T} \sum_{t=1}^T Acc_t$. As the ultimate goal of CIL is to maintain knowledge after all tasks, $Acc_{\text{Last}}$ is considered as the most critical metric, while $Acc_{\text{Avg}}$ reflects stability throughout training. All methods use the same seeds and PFMs for a fair comparison. 
The implementation is available at \url{https://github.com/CUHK-BMEAI/CRCL/}.

\subsection{Main Results}
% \subsubsection{Comparison with SOTA Methods}
Table \ref{tab_main} presents the results of various methods across five medical datasets. Joint training serves as the upper bound, representing the ideal scenario with simultaneous access to all data, while continual PFM finetuning acts as the lower bound, highlighting severe forgetting. The compared methods can be categorized into traditional CIL (e.g., FOSTER \cite{wang2022foster}, iCaRL \cite{rebuffi2017icarl} and DER \cite{yan2021DER}) and PFM-based CIL (e.g., ACL \cite{AdapterCL2023}, L2P \cite{L2P2022}, DualPrompt \cite{DualP2022}, CODAPrompt \cite{CODAP2023}, LAE \cite{gao2023LAE}, ADAM-Adapter \cite{zhou2024adam}, SLCA \cite{zhang2023slca}, EASE \cite{zhou2024ease}, SSIAT \cite{tan2024shift}, and MOS \cite{sun2024mos}). While traditional top-performing CIL methods achieve competitive results on simpler datasets (e.g., Colon and Blood), their reliance on computationally demanding tuning and data replay raises practical concerns. PFM-based CIL methods utilize PET to enable adaptability while reducing computational overhead. Yet, existing methods exhibit notable performance variability with suboptimal results. As shown, ACL \cite{AdapterCL2023} achieves competitive performance but depends on replaying prior data. ADAM-Adapter \cite{zhou2024adam}, leveraging prototypes and first-session adaptation, enhances efficiency but lacks continual adaptation. SSIAT \cite{tan2024shift}, incorporating semantic shift estimation, improves feature alignment but struggles with Colon, Skin8 and MedMNIST. The recent MOS \cite{sun2024mos} utilizes adapter merging and a self-refined adapter retrieval mechanism, yet it still yields suboptimal performance. Meanwhile, prompt-based methods (L2P \cite{L2P2022}, DualPrompt \cite{DualP2022} and CODAPrompt \cite{CODAP2023}) show overall weaker performance, likely due to the limited expressiveness of soft prompts in handling complex distributions of medical images. We also present the incremental performance trend of various methods in Fig. \ref{fig_results}, observing that most methods struggle with maintaining stability across sessions, with significant performance drops in later stages. Encouragingly, Bi-CRCL consistently attains superior performance compared with existing replay-free PFM-based approaches across all datasets. As shown in Fig. \ref{fig_results}, Bi-CRCL maintains more stable performance throughout incremental sessions, indicating effective mitigation of catastrophic forgetting. These results suggest that Bi-CRCL achieves a balanced trade-off between stability and plasticity, enabling robust continual adaptation to evolving medical data without relying on replay.

\subsection{In-depth Analysis}

\subsubsection{Effectiveness on General-domain and Medical-specific PFMs}
\label{sec:medptm}
While our main experiments adopt the general-domain PFM (ViT-B/16 pretrained on ImageNet-21K and fine-tuned on ImageNet-1K) by default, an important question is whether Bi-CRCL can also generalize to medical-domain PFMs. To examine this, we further evaluate Bi-CRCL on two recent representative medical PFMs, BiomedCLIP~\cite{zhang2023biomedclip} and RAD-DINO~\cite{rad-dino2025}, as summarized in Table~\ref{tab:ours_medptm}. BiomedCLIP is pretrained through contrastive vision-language alignment on PMC-15M, a large-scale corpus containing approximately 15 million biomedical image-caption pairs spanning diverse modalities such as radiography, microscopy, histopathology, and biomedical illustrations. In contrast, RAD-DINO follows a self-supervised DINOv2 paradigm trained on large collections of chest X-ray images without text supervision, aiming to learn radiology-oriented visual representations.
Interestingly, our results show that these medical PFMs do not consistently outperform their general-domain counterpart. This indicates that domain specialization does not necessarily yield stronger or more transferable representations across heterogeneous medical datasets. Nonetheless, Bi-CRCL demonstrates stable improvements across both general-domain and medical-domain PFMs, confirming its robustness to backbone variations. Moreover, we observe complementary strengths: when applied to medical PFMs, Bi-CRCL effectively leverages domain-specific priors; when applied to general-domain PFMs, it enhances adaptation by recalibrating generalized features.
Overall, these findings suggest that Bi-CRCL serves as a unifying continual learning framework capable of bridging general- and medical-domain PFMs. It is important to note, however, that all approaches fundamentally rely on the quality of the pretrained feature representations. Continual learning strategies cannot fully compensate for weak or poorly aligned PFMs, and the ultimate performance ceiling remains constrained by the representational strength of the underlying model. We therefore advocate a co-evolutionary perspective, where continual learning strategies and foundation model pretraining paradigms advance synergistically to drive future progress in medical CIL.

\begin{table*}[!t]
\small
\centering
\caption{Performance comparison using different PFMs (medical-specific vs. general-domain) across five benchmark datasets. The best results of each PFM are \textbf{bolded}.}
\resizebox{0.95\textwidth}{!}{%
\begin{tabular}{l|l|cc|cc|cc|cc|cc}
\Xhline{1pt}
%\multicolumn{10}{c}{\textbf{Medical Domain}} \\
%\midrule
\bf \multirow{2}{*}{\textbf{PFM}} & \bf\multirow{2}{*}{\textbf{Method}} & \multicolumn{2}{c|}{\textbf{Colon}} & \multicolumn{2}{c|}{\textbf{Blood}} & \multicolumn{2}{c|}{\textbf{Skin8}} & \multicolumn{2}{c|}{\textbf{MedMNIST-Sub}} & \multicolumn{2}{c}{\textbf{COVID (CT\&X-rays)}} \\
& & $Acc_{\text{Avg}}$ (\%) & $Acc_{\text{Last}}$ (\%) & $Acc_{\text{Avg}}$ (\%) & $Acc_{\text{Last}}$ (\%) & $Acc_{\text{Avg}}$ (\%) & $Acc_{\text{Last}}$ (\%) & $Acc_{\text{Avg}}$ (\%) & $Acc_{\text{Last}}$ (\%)& $Acc_{\text{Avg}}$ (\%) & $Acc_{\text{Last}}$ (\%) \\
\hline
\multirow{6}{*}{RAD-DINO  \cite{rad-dino2025}} 
& Joint Training & - & 86.34 & - & 65.83 & - & 42.20 & - & 56.27 & - & 70.80 \\
& Finetuning & 38.06 & 11.57 & 44.84 & 36.57 & 31.76 & 15.32 & 48.80 & 33.53 & 24.39 & 6.25 \\
& SimpleCIL \cite{zhou2024adam}& 53.44 & 46.67 & 43.14 & 25.29 & 34.82 & 21.56 & 48.62 & 38.01 & 57.09 & 46.60 \\
& ADAM-Adapter \cite{zhou2024adam} & 54.92 & 48.58 & 42.50 & 24.55 & 35.40 & 21.99 & 49.91 & 38.06 & 55.51 & 39.46 \\
& MOS \cite{sun2024mos} & 24.52 & 8.77 & 31.68 & 9.08 & 24.09 & 5.53 & 22.95 & 8.23 & 21.28 & 11.01 \\
& Bi-CRCL (ours) & \textbf{85.07} & \textbf{74.28} & \textbf{66.28} & \textbf{46.98} & \textbf{48.93} & \textbf{32.77} & \textbf{68.69} & \textbf{53.39} & \textbf{81.93} & \textbf{70.80} \\ \hdashline
\multirow{6}{*}{BiomedCLIP \cite{zhang2023biomedclip}} 
& Joint Training & - & 97.50 & - & 96.38 & - & 55.74 & - & 66.68 & - & 89.65 \\
& Finetuning & 66.14 & 51.41 & 48.66 & 26.36 & 34.46 & 15.18 & 46.06 & 24.21 & 21.66 & 7.28 \\
& SimpleCIL \cite{zhou2024adam}& 93.28 & 89.88 & 85.86 & 85.17 & 57.43 & 40.99 & 69.86 & 53.21 & 83.51 & 65.30\\
& ADAM-Adapter \cite{zhou2024adam} & 93.49 & 90.19 & 79.12 & 78.74 & 54.79 & 38.87 & 70.88 & 53.48 & 83.38 & 65.21 \\
& MOS \cite{sun2024mos} & 96.03
& 94.64 & 90.33 & 89.58 & 67.48 & 51.91 & 78.67 & 63.96 & 90.62 & 73.97 \\
& Bi-CRCL (ours) & \textbf{98.53} & \textbf{97.57} & \textbf{96.97} & \textbf{96.45} & \textbf{69.20} & \textbf{53.62} & \textbf{82.16} & \textbf{66.56} & \textbf{96.04} & \textbf{88.06} \\ \hdashline
\multirow{6}{*}{ViT-B/16-IN1K \cite{alexey2020image}} 
& Joint Training & - & 99.98 & - & 99.61 & - & 67.73 & - & 73.61 & - & 92.43 \\
& Finetuning & 38.65 & 10.43 & 37.47 & 14.54 & 39.62 & 17.87 & 29.18 & 5.66  & 26.55 & 10.54 \\
& SimpleCIL \cite{zhou2024adam} & 90.10 & 85.41 & 83.85 & 79.79 & 56.61 & 38.30 & 68.07 & 50.63 & 75.84 & 57.37 \\ 
& ADAM-Adapter \cite{zhou2024adam} & 86.01 & 78.00 & 88.09 & 83.52 & 59.82 & 41.84 & 70.97 & 53.11 & 79.19 & 61.10 \\ 
& MOS \cite{sun2024mos} & 91.46 & 87.60 & 92.53 & 90.18 & 68.54 & 51.77 & 74.59 & 51.80 & 89.96 & 80.60 \\ 
& Bi-CRCL (ours) & \textbf{99.12} & \textbf{98.51} & \textbf{98.08} & \textbf{97.56}  & \textbf{74.59} & \textbf{61.99} & \textbf{84.22} & \textbf{69.71} & \textbf{96.12} & \textbf{88.15} \\
\Xhline{1pt}
\end{tabular}%
}
% \vspace{0.3cm}
\begin{minipage}{0.93\textwidth}
\tiny
\textit{Note:} SimpleCIL is a feature-quality check baseline presented in \cite{zhou2024adam}, using a frozen PFM with a nearest-class-mean classifier.
\end{minipage}
% \vspace{-0.3cm}
\label{tab:ours_medptm}
\end{table*}

\subsubsection{Efficacy of Each Component}
To better understand Bi-CRCL, we present an ablation study (Table \ref{tab:ablation_without}) on three representative and relatively challenging datasets (Skin8, COVID and MedMNIST-Sub), covering natural, radiological and heterogeneous multi-organ domains. The ablation results consistently validate the contribution of each module. Excluding initialized adaptation (Abla-1) leads to a notable performance drop, indicating that while general-domain PFMs offer generalizable features, domain adaptation remains essential for downstream medical applications. Removing bidirectional knowledge interaction (Abla-2) degrades performance, underscoring its key role in enhancing previous one-way knowledge transfer (CRCL) and initializing new learning with consolidated knowledge, thereby providing a warm start that enables new knowledge to build upon long-term consolidated information. Removing conservative learner inference (Abla-3) causes a moderate decline, highlighting the role of the neocortex in decision-making based on consolidated experience. Excluding radical learner inference (Abla-4) results in a smaller drop, underscoring the complementary roles of the two learners and suggesting that the absence of either may lead to overfitting to earlier or later tasks, compromising overall robustness. These results confirm that each Bi-CRCL component contributes to performance improvement, collectively fostering an effective balance between stability and plasticity, while preserving PFM generalizability and enabling downstream adaptability.

\begin{table}[t]
\centering
\caption{Ablation analysis on three representative and challenging datasets (Skin8, COVID and MedMNIST-Sub). The best and second-best results are \textbf{bolded} and \underline{underlined}, respectively.}
\label{tab:ablation_without}
\scalebox{0.62}{
\begin{tabular}{p{2cm}<{\centering}|p{1.5cm}<{\centering}|p{5cm}<{\centering}|p{1.5cm}<{\centering}|p{1.5cm}<{\centering}}
\Xhline{1pt}
\textbf{Dataset} & \textbf{Setting} & \multicolumn{1}{c|}{\textbf{Exclusion}} & $Acc_{\text{Avg}}$ (\%) & \multicolumn{1}{c}{$Acc_{\text{Last}}$ (\%)} \\\hline
\multirow{5}{*}{Skin8} 
&Bi-CRCL         & None                                     & \textbf{74.59}       & \textbf{61.99} \\
&Abla-1          & Initialized Adaptation                         & 70.96       & 59.69          \\
&Abla-2          & Bidirectional Knowledge Interaction      & 74.83 & 61.42       \\
&Abla-3          & Conservative Learner Inference          & 73.82 & 60.43      \\
&Abla-4          & Radical Learner Inference          & \underline{74.49} & \underline{61.84}       \\
\hline
\multirow{5}{*}{COVID} 
& Bi-CRCL  & None                                     & \textbf{96.12}       & \textbf{88.15}          \\
& Abla-1   & Initialized Adaptation                   & 95.66       & 84.24          \\
& Abla-2   & Bidirectional Knowledge Interaction      & 95.46       & 84.70          \\
& Abla-3   & Conservative Learner Inference           & 95.68       & 86.15          \\
& Abla-4   & Radical Learner Inference                & \underline{95.98}       & \underline{87.94}          \\\hline
\multirow{5}{*}{MedMNIST-Sub} 
& Bi-CRCL  & None                                     & \underline{84.22}       & \textbf{69.71}          \\
& Abla-1   & Initialized Adaptation                   & 77.56       & 55.35          \\
& Abla-2   & Bidirectional Knowledge Interaction      & 83.20       & 65.43          \\
& Abla-3   & Conservative Learner Inference           & \textbf{84.69}       & 68.51          \\
& Abla-4   & Radical Learner Inference                & 84.04      & \underline{68.88}          \\
\Xhline{1pt} 
\end{tabular}}
\vspace{-0.3cm}
\end{table}

\subsubsection{Sensitivity to Task Order}
\label{sec:data_order}
To assess the sensitivity of each method to task sequencing, we conduct experiments using both a predefined randomly shuffled task order and its reversed counterpart, ensuring that all baselines are evaluated under identical conditions. As shown in Table~\ref{tab:reverse_order}, the performance of Bi-CRCL across five benchmark medical datasets remains highly stable, indicating low sensitivity to task order and minimal dependence on explicit task boundaries.
For comprehensive comparison, we include recent competitive baselines such as MOS~\cite{sun2024mos} and SSIAT~\cite{tan2024shift}, as well as the SimpleCIL prototypical classifier~\cite{zhou2024adam} and standard fine-tuning. These methods show notable performance degradation under the reversed task order, whereas Bi-CRCL consistently outperforms them while maintaining stability across all settings.

\begin{table*}[!h]
\small
\centering
\caption{Performance comparison across five benchmark datasets using a reversed task order. The best and second-best results are \textbf{bolded} and \underline{underlined}, respectively.}
\resizebox{0.95\textwidth}{!}{%
\begin{tabular}{cc|cc|cc|cc|cc|cc}
\Xhline{1pt}
%\midrule
\bf \multirow{2}{*}{\textbf{Method}} & \bf \multirow{2}{*}{\textbf{Order}} & \multicolumn{2}{c|}{\textbf{Colon}} & \multicolumn{2}{c|}{\textbf{Blood}} & \multicolumn{2}{c|}{\textbf{Skin8}} & \multicolumn{2}{c|}{\textbf{MedMNIST-Sub}} & \multicolumn{2}{c}{\textbf{COVID (CT\&X-rays)}} \\
& & $Acc_{\text{Avg}}$ (\%) & $Acc_{\text{Last}}$ (\%) & $Acc_{\text{Avg}}$ (\%) & $Acc_{\text{Last}}$ (\%) & $Acc_{\text{Avg}}$ (\%) & $Acc_{\text{Last}}$ (\%) & $Acc_{\text{Avg}}$ (\%) & $Acc_{\text{Last}}$ (\%)& $Acc_{\text{Avg}}$ (\%) & $Acc_{\text{Last}}$ (\%) \\
\hline
Finetuning  & - & 38.65 & 10.43 & 37.47 & 14.54 & 39.62 & 17.87 & 29.18 & 5.66  & 26.55 & 10.54 \\
Finetuning & reversed & 47.80 & 26.52 & 49.67 & 22.97 & 25.32 & 14.75 & 31.13 & 18.45 & 34.47 & 11.19 \\\hdashline
SimpleCIL \cite{zhou2024adam} & - & 90.10 & 85.41 & 83.85 & 79.79 & 56.61 & 38.30 & 68.07 & 50.63 & 75.84 & 57.37 \\
SimpleCIL \cite{zhou2024adam} & reversed & 90.17 & 85.91 & 89.54 & 79.85 & 53.79 & 38.30 & 49.58 & 50.44 & 77.88 & 66.79 \\\hdashline
SSIAT \cite{tan2024shift} & - & 75.36 & 66.34 & 86.00 & 84.63 & 60.46 & 41.99 & 59.43 & 25.79 & 72.00 & 60.17 \\
SSIAT \cite{tan2024shift} & reversed & 83.73 & 70.17 & 83.07 & 52.28 & 62.51 & 47.94 & 40.84 & 24.06 & 74.09 & 64.18 \\\hdashline
MOS \cite{sun2024mos} & - & 91.46 & 87.60 & 89.27 & 87.39 & 63.80 & 45.82 & 74.59 & 51.80 & 81.07 & 68.19 \\
MOS \cite{sun2024mos}& reversed & 93.98 & 91.94 & 92.03 & 85.04 & 65.36 & 49.65 & 62.32 & 62.69 & 71.48 & 71.36 \\\hdashline
Bi-CRCL (ours) & - & \underline{99.12} & \textbf{98.51} &  \underline{98.08} & \textbf{97.56}  & \underline{74.59} & \underline{61.99} & \textbf{84.22} & \underline{69.71} & \textbf{96.12} & \textbf{88.15} \\
Bi-CRCL (ours) & reversed & \textbf{99.21} & \underline{98.48} & \textbf{98.97} & \underline{97.49} & \textbf{75.40} & \textbf{62.13} &  \underline{80.08} & \textbf{70.25} &  \underline{93.85} & \underline{85.82} \\
%\bottomrule
\Xhline{1pt}
\end{tabular}%
}
\label{tab:reverse_order}
\end{table*}

\subsubsection{Robustness to the Number of Tasks}
\label{sec:data_split}
We conduct an empirical study on the MedMNIST-Sub dataset, which contains a relatively large number of classes, to investigate how the number of tasks affects long-term continual learning performance. MedMNIST-Sub is not only the dataset with the most classes in our benchmark but also one of the most diverse and challenging. Therefore, it serves as a suitable testbed for this analysis. Specifically, we divide the classes of MedMNIST-Sub into eight tasks, compared with the standard four-task setting. As shown in Table~\ref{tab:task_num}, increasing the task number has only a marginal impact on overall performance, indicating that the task-split strategy is not a primary factor influencing long-term continual learning capability.

\begin{table}[t]
\centering
\caption{Effect of task-split granularity on Bi-CRCL performance (MedMNIST-Sub dataset).}
\label{tab:task_num}
\scalebox{0.65}{
\begin{tabular}{p{3cm}<{\centering}|p{2.5cm}
<{\centering}|p{2.5cm}<{\centering}|p{2.5cm}<{\centering}}
\Xhline{1pt}
\textbf{Method} & \textbf{Task Number} & $Acc_{\text{Avg}}$ (\%) & $Acc_{\text{Last}}$ (\%) \\\hline
SimpleCIL \cite{zhou2024adam} & 4 & 68.07 & 50.63 \\
SimpleCIL \cite{zhou2024adam} & 8 & 51.41 & 50.44 \\\hdashline
MOS \cite{sun2024mos} & 4 & 74.59 & 51.80 \\
MOS \cite{sun2024mos} & 8 & 80.96 & 61.83 \\\hdashline
Bi-CRCL (ours) & 4 & 84.22  & 69.71 \\
Bi-CRCL (ours)  & 8 & 86.48 & 70.28 \\
\Xhline{1pt}                               
\end{tabular}}
% \vspace{-0.3cm}
\end{table}

\subsubsection{Necessity of Memory Replay}
\label{sec:replay}
% 4. Discussion on whether adding representative samples for replay can further enhance the performance - CRCL under replay-free and replay-available performances
To examine the necessity of replay-based techniques within our framework, we integrate a memory buffer following iCaRL~\cite{rebuffi2017icarl} and apply the widely used herding-based exemplar selection strategy~\cite{welling2009herding}. This strategy selects class exemplars that best approximate the class mean in feature space, aiming to preserve representative features from past tasks. We adopt the original iCaRL configuration with a buffer of up to 2,000 samples, and reduce it to 400 samples for datasets such as Skin8 and COVID due to class imbalance, where certain categories lack sufficient examples to support larger buffers. We then evaluate Bi-CRCL under both replay-free and replay-enabled settings across multiple medical datasets. Interestingly, as illustrated in Table \ref{tab:replay_result}, introducing a replay memory buffer does not consistently improve performance and even results in marginal decreases in $Acc_{\text{Last}}$ across most datasets. This suggests that overfitting to stored samples may offset the potential advantages of replay. More importantly, these results indicate that Bi-CRCL, through the complementary interaction between its conservative and radical learners, already preserves critical class-specific knowledge and mitigates forgetting effectively even without access to replay data. These findings open up a valuable insight for future PFM-based medical CIL research: with a well-designed and robust framework, strong continual learning performance can be achieved without heavy reliance on replay, which is especially relevant in privacy-sensitive clinical settings.

\begin{table}[t]
\centering
\caption{Empirical analysis of memory replay in Bi-CRCL.}
\label{tab:replay_result}
\scalebox{0.65}{
\begin{tabular}{p{3cm}<{\centering}|p{1.5cm}<{\centering}|p{2.5cm}<{\centering}|p{2.5cm}<{\centering}}
\Xhline{1pt}
\textbf{Dataset} & \textbf{Replay} & $Acc_{\text{Avg}}$ (\%) & \multicolumn{1}{c}{$Acc_{\text{Last}}$ (\%)} \\\hline
\multirow{2}{*}{Colon} 
& $\checkmark$           & 99.06    & 98.43 \\
& $\times$                & 99.12   & 98.51 \\
\hdashline
\multirow{2}{*}{Blood} 
& $\checkmark$           & 98.06       & 97.53 \\
& $\times$                & 98.08   & 97.56 \\
\hdashline
\multirow{2}{*}{Skin8} 
& $\checkmark$           & 74.56       & 61.99 \\
& $\times$                & 74.59   & 61.99 \\
\hdashline
\multirow{2}{*}{MedMNIST-Sub} 
& $\checkmark$           & 84.16      & 69.32 \\
& $\times$                & 84.22   & 69.71 \\
\hdashline
\multirow{2}{*}{COVID (CT\&X-ray)} 
& $\checkmark$           & 95.81   & 86.10 \\
& $\times$                & 96.12   & 88.15 \\
\Xhline{1pt}                               
\end{tabular}}
% \vspace{-0.3cm}
\end{table}

%\subsubsection{Class Prototype Feature Analysis}
%\label{sec:feature_map}
%draw what[Placeholder]

\subsubsection{Generalizability on Inter-dataset Continual Learning}
We establish a new inter-dataset benchmark using the aforementioned five datasets to evaluate continual learning methods under more challenging medical scenarios. Prior analyses are typically constructed by splitting classes within a single dataset \cite{kather2019predicting,acevedo2019recognition,tschandl2018ham10000}, where typically all images and classes belong to the same domain (e.g., skin, colon, or blood). Here, this benchmark spans multiple medical domains, each requiring distinct expert knowledge and interpretation. This setup introduces greater heterogeneity and poses a more comprehensive challenge for model evaluation. As shown in Table \ref{tab:interdataset}, methods like SLCA \cite{zhang2023slca} and SSIAT \cite{tan2024shift} suffer from inter-task forgetting and exhibit a noticeable performance drop compared to prior standard intra-dataset experiments. SimpleCIL~\cite{zhou2024adam}, ADAM-Adapter~\cite{zhou2024adam} and MOS~\cite{sun2024mos} show limited generalization. Remarkably, our Bi-CRCL maintains stable performance and clearly outperforms these baselines in both $Acc_{\text{Avg}}$ and $Acc_{\text{Last}}$, even interestingly surpassing the upper bound of Joint Training, which has access to all task data simultaneously. We attribute this to the tendency of Joint Training to overfit domain-specific biases, whereas Bi-CRCL incrementally consolidates knowledge through bidirectional interaction, fostering invariant representations and stronger cross-domain generalization.

\begin{table}[t]
\centering
\caption{Performance comparison on inter-dataset continual learning. The best results are \textbf{bolded}.}
\label{tab:interdataset}
\scalebox{0.75}{
\begin{tabular}{p{3cm}<{\centering}|p{2.5cm}<{\centering}|p{2.5cm}<{\centering}}
\Xhline{1pt}
\textbf{Method} & $Acc_{\text{Avg}}$ (\%) & $Acc_{\text{Last}}$ (\%) \\\hline
Joint Training & - & 71.65 \\
SimpleCIL \cite{zhou2024adam} & 57.50 & 55.41 \\
ADAM-Adapter \cite{zhou2024adam} & 63.18 & 55.52 \\
SLCA \cite{zhang2023slca} & 38.34 & 33.72 \\
SSIAT \cite{tan2024shift} & 53.97 & 28.41  \\
MOS \cite{sun2024mos} & 70.66 & 62.05 \\
% SAFE \cite{zhao2024safe} & 73.77 & 71.66 \\
Bi-CRCL (ours) & \textbf{88.62}       & \textbf{75.01} \\
\Xhline{1pt}                               
\end{tabular}}
\vspace{-0.3cm}
\end{table}

\section{Discussion}
\label{sec_discuss}
The proposed Bidirectional Conservative-Radical Complementary Learning (Bi-CRCL) framework achieves consistent and substantial performance gains across diverse medical imaging datasets and evaluation settings. On standard continual learning benchmarks, Bi-CRCL surpasses recent state-of-the-art replay-free and replay-based methods in both last-session and average accuracy, reflecting an effective stability-plasticity balance throughout the continual learning process. These results validate the core principle of Bi-CRCL: a bidirectional interaction between a stability-oriented conservative learner and a plasticity-driven radical learner, where dynamic knowledge exchange enables continual adaptation without catastrophic forgetting. Visualization of incremental performance trajectories further highlights Bi-CRCL’s ability to maintain generalization while progressively integrating new task knowledge. Even under more challenging inter-dataset conditions involving severe distribution shifts, Bi-CRCL sustains robust performance and significantly outperforms competitive methods. Additional evaluations across varying task orders and task quantities further confirm its resilience and scalability.

Our empirical analysis of different PFMs shows that medical-domain PFMs often underperform relative to general-domain PFMs, indicating that domain specialization does not necessarily yield more transferable or discriminative representations. This observation reinforces a central motivation of this work—the need for replay-free continual adaptation strategies that can effectively bridge powerful general-domain PFMs to evolving medical tasks while maintaining their broad generalization capability. Nevertheless, the development of higher-quality and more diverse medical PFMs remains an important direction for future research. In addition, experiments incorporating memory replay produce negligible improvements, suggesting that the proposed dual-learner design already captures essential class-specific information and maintains a balanced trade-off between generalization and adaptation. Together, these results highlight Bi-CRCL as a scalable, robust and replay-free continual learning framework well suited for medical image analysis.

Despite these encouraging results, certain limitations remain. First, while the dual-learner architecture balances generalization and adaptation effectively, it introduces a modest computational overhead due to maintaining two learners in parallel. Second, performance may degrade when new tasks have limited training samples, making it difficult to compute representative class prototypes. Future work could explore more extreme yet realistic scenarios, such as few-shot and data-imbalanced continual learning settings, as well as extensions to other medical-domain tasks such as continual segmentation.

In summary, Bi-CRCL achieves a balanced integration of continual learning stability and adaptability. Its replay-free design, coupled with the bidirectional conservative-radical interaction, makes it particularly well-suited for real-world clinical diagnostic workflows where scalability, robustness, privacy, and domain adaptability are essential. The framework’s versatility in handling diverse domains and evolving disease spectra underscores its promise as a foundation for next-generation continual disease diagnosis AI systems in practical clinical settings.
% \subsection{Impacts}

% \subsection{Future Works}

\section{Conclusion}
In this work, we introduced Bidirectional Conservative-Radical Complementary Learning (Bi-CRCL), a replay-free framework for class-incremental medical image analysis. The core of Bi-CRCL is the continual bidirectional knowledge interaction between a stability-oriented conservative learner and a plasticity-driven radical learner, where dynamic knowledge exchange and consolidation balance the generalization of foundation models with downstream adaptation to yield robust task-agnostic predictions. Extensive and challenging evaluations across five medical datasets demonstrated that Bi-CRCL outperformed recent top-performing methods. By bridging general-domain pretrained foundation models with clinical demands, Bi-CRCL advances scalable and lifelong diagnostic systems that adapt to evolving disease diversity.

% or


% \appendices

% % you can choose not to have a title for an appendix
% % if you want by leaving the argument blank
% \section{}

% % use section* for acknowledgment
% \ifCLASSOPTIONcompsoc
%   % The Computer Society usually uses the plural form
%   \section*{Acknowledgments}
% \else
%   % regular IEEE prefers the singular form
% \fi
% This research was supported by General Research Fund from Research Grant Council of Hong Kong (No. 14205419).

% \ifCLASSOPTIONcaptionsoff
%   \newpage
% \fi

\bibliographystyle{IEEEtran.bst}
\bibliography{IEEEexample.bib} 

%\begin{minipage}[t]{0.457\textwidth}
% \vskip -1\baselineskip plus -1fil
%\vspace*{-5\baselineskip}

% \end{minipage}
%\enlargethispage{-9.5cm}
% \vskip -1\baselineskip plus -1fil

% If someone has no photo:
% \begin{IEEEbiographynophoto}{Fourth D. Author}
% is with GHI Institute. His/Her interests include X, Y, and Z.

\end{document}